\documentclass[10pt,twocolumn,letterpaper]{article}
\usepackage{cvpr}
\usepackage{array}
\usepackage{multirow}
\usepackage{pifont}
\newcommand{\xmark}{\ding{55}}

\definecolor{cvprblue}{rgb}{0.21,0.49,0.74}
\usepackage[pagebackref,breaklinks,colorlinks,allcolors=cvprblue]{hyperref}
\hypersetup{
  pdftitle={LC-ISP Benchmark: Learning Advanced ISP with Live and Capture Streams},
  pdfauthor={Kai Chen, Junyu Lou, Jinhua Zhang, Weiyi You, Shuhang Gu}
}
\def\confName{CVPR}
\def\confYear{2027}
\title{LC-ISP Benchmark: Learning Advanced ISP with Live and Capture Streams}
\author{%
Kai Chen \quad Junyu Lou \quad Jinhua Zhang \quad Weiyi You \quad Shuhang Gu\thanks{Corresponding author.}\\
University of Electronic Science and Technology of China\\
{\tt\small \{kkchen318, shuhanggu\}@gmail.com}\\
{\tt\small \url{https://github.com/LabShuHangGU/LC-ISP}}
}

\begin{document}
\maketitle
\begin{abstract}
High-quality paired RAW-to-sRGB datasets are essential for data-driven ISP, yet existing datasets often rely on cross-device collection, causing misalignment and failing to faithfully reflect real capture conditions. We therefore introduce \textbf{LC-ISP}, a large-scale benchmark built from the intrinsic gap between a camera's live preview and offline capture streams. LC-ISP contains 53.9k spatially aligned groups, each comprising live RAW/live sRGB and multi-frame fused RGB/capture sRGB, and covers day, night, indoor, outdoor, and HDR scenes. It also provides capture-stream intermediates for modular supervision. We further develop a compact two-stage baseline that uses these intermediates as auxiliary targets. Under a 2.5-GFLOP budget for 448$\times$448 RAW inputs, it outperforms the strongest single-frame baseline by over 2 dB PSNR while improving texture, denoising, and color reproduction. LC-ISP supports realistic benchmarking, the design of supervision strategies, and the development of lightweight ISP models.

\end{abstract}
\begin{figure*}[t]
  \centering
  \includegraphics[width=\textwidth]{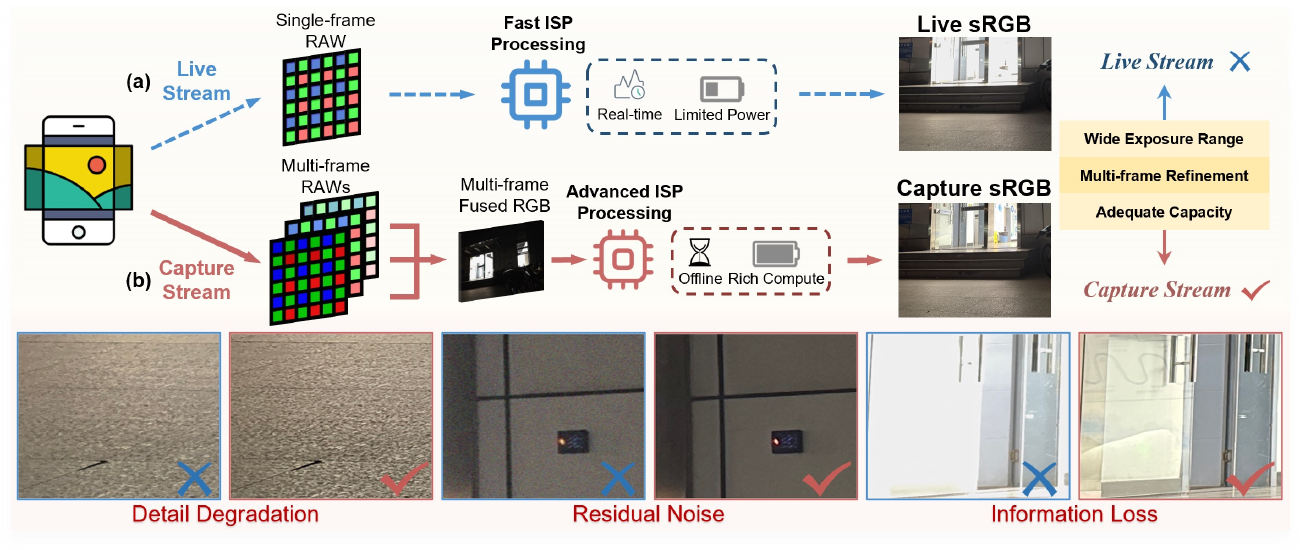}
  \vspace{-15pt}
  \caption{
    Motivation: Our method leverages the inherent differences between preview live streams and offline capture streams to construct a large-scale benchmark. (a) Live stream: A lightweight, single-frame pipeline for real-time sRGB rendering. (b) Capture stream: Multi-frame fusion with exposure bracketing to reconstruct high-quality fused RGB and sRGB outputs. The bottom row compares their visual results, highlighting the quality gap between the two streams.
  }
  \label{problem}
  \vspace{-10pt}
\end{figure*}

\section{Introduction}
Image Signal Processing (ISP) is a fundamental component of modern imaging systems, responsible for converting RAW sensor measurements into display-ready images for visualization, composition, and downstream tasks.
Driven by rapid advancements in deep learning, neural-network-based ISPs have received widespread attention. A key factor in this progress is the availability of paired ISP datasets, captured with imaging hardware under favorable acquisition conditions \cite{ignatov2020replacingmobilecameraisp,ignatov2021learnedsmartphoneispmobile}. Leveraging these datasets, recent works have advanced RAW-to-sRGB mappings by proposing powerful network architectures \cite{ignatov2020aim2020challengelearned,zhang2021learningrawtosrgbmappingsinaccurately,dai2020awnetattentivewaveletnetwork}.

Although previous paired datasets have played a crucial and foundational role in advancing the field of ISP, they continue to face significant challenges when applied in real-world deployment scenarios. The first major issue is the misalignment problem, where datasets are collected using different devices, leading to inevitable misalignment between the paired images. The second challenge is that these datasets are fundamentally constrained by the restricted dynamic range of a single-frame sensor exposure, which makes it difficult to provide high-quality supervision for training ISP models with both robustness and generalization capability.

To address these challenges and expand the available supervisory signals without introducing cross-device mismatch, we propose a practical data-construction strategy that leverages the natural differences between on-device live streams and offline capture streams.
Concretely, many edge devices such as smartphones operate two concurrent imaging streams. One is a live stream, processed by a low-latency on-chip ISP, which runs at a high frame rate to provide a live view on the device screen. The other is a capture stream that, upon shutter press, can fuse multiple RAW exposures with a larger computational budget to produce high-bit, high-fidelity outputs. As illustrated in Figure \ref{problem}, the live stream is constrained by tight latency budgets and thus relies on lightweight single-frame processing; the capture stream, by contrast, can employ multi-frame fusion and heavier neural processing to recover higher SNR, wider dynamic range, and finer texture. Due to the disparity between single-frame and multi-frame inputs, these operational differences create significant information gaps between the two streams, leading to inherent quality differences in detail, noise level, color fidelity, and dynamic range. We are motivated by the observation that these same-device, stream-level differences provide a practical, well-aligned source of high-quality supervision for improving real-time ISP models while avoiding misalignment caused by acquisition with different devices.

To that end, we introduce \textbf{LC-ISP}, a live-and-capture stream ISP benchmark collected on commodity smartphone hardware (iPhone 16 Pro). LC-ISP provides spatially aligned single-frame RAW inputs and multi-frame fused capture sRGB, with each example consisting of four images: the live stream’s RAW and sRGB frames, and the capture stream’s multi-frame fused RGB and capture sRGB. It covers a variety of scenes and lighting conditions (day, night, indoor, outdoor, HDR) and contains approximately 53.9k aligned pairs. The benchmark is designed with two main objectives: (1) to better promote the research of AI-ISP methods, and (2) to facilitate the narrowing of the quality gap between live ISPs and offline capture ISPs.
Another key advantage of LC-ISP over existing AI-ISP benchmarks is its ability to collect intermediate processing data. This capability not only supports functional ISP research but also enables the exploration of more flexible supervision methods. By providing intermediate outputs, LC-ISP allows for more comprehensive supervision during training, fostering better model performance. Leveraging LC-ISP, we develop a compact two-stage baseline architecture, Dual-Stage ISP, and demonstrate that, under typical compute budgets representative of real-time live ISPs, introducing intermediate supervision from the capture stream outputs significantly improves the quality of single-frame RAW-to-sRGB mapping. This results in sharper textures, more robust noise suppression, and more accurate color reproduction. In particular, on 448$\times$448 RAW inputs and under the same computational budget of 2.5 GFLOPs, our approach outperforms the best single-frame baseline by more than 2 dB in terms of PSNR, achieving a noticeable enhancement in image fidelity.

Our contributions are summarized as follows:
\begin{itemize}
    \item We propose a practical dataset construction strategy that leverages intrinsic differences between on-device live and capture streams to produce high-quality, well-aligned training pairs for AI-ISP research.
    \item We introduce LC-ISP, a live-and-capture stream benchmark recorded on smartphone hardware that covers diverse scenes and provides both final sRGB captures and intermediate fused outputs to support flexible research.
    \item We design a lightweight two-stage baseline, Dual-Stage ISP, and demonstrate through extensive experiments that incorporating intermediate supervision from capture stream can markedly reduce the live–capture quality gap under realistic mobile compute constraints.
\end{itemize}

\section{Related work}
\textbf{ISP datasets}
Learning-based image signal processing methods often rely on real-world datasets. Early works mainly focus on image denoising, image enhancement and high dynamic range imaging. The Darmstadt Noise Dataset (DND) \cite{dnd} provides carefully calibrated noisy/clean pairs acquired with consumer cameras of different sensor sizes, and has become a standard benchmark for evaluating denoising algorithms under realistic ISP noise characteristics. To better promote the development of low-light denoising, the SID dataset \cite{chen2018learningdark} collects paired short-exposure RAW images and long-exposure references in extreme low-light conditions, and demonstrates that a neural network operating directly on RAW sensor data can effectively replace much of the conventional ISP pipeline for night-time imaging. For HDR reconstruction of dynamic scenes, Kalantari \etal \cite{kalantari_hdr} introduce a widely used dataset of multi-exposure LDR triplets with ground-truth HDR images. The MIT-Adobe FiveK dataset \cite{mit5k} contains 5{,}000 RAW images with expert-retouched versions for tone and color enhancement. Live2K \cite{lou2026perceivingbettermomentscover} further provides 2{,}042 real-world Live Photos to support reference-guided restoration of both spatial and color quality in the corresponding cover frames.

To promote end-to-end learning of the ISP pipeline, more recent works construct datasets that explicitly pair RAW sensor measurements with high-quality sRGB renderings. The DPED dataset \cite{dped} aligns smartphone images or RAW data with DSLR photographs, allowing models to learn a mapping from mobile images to DSLR-quality outputs and effectively viewing the ISP as a learnable RAW-to-sRGB function. Building on this idea, subsequent datasets such as ZRR \cite{ignatov2020replacingmobilecameraisp} and MAI \cite{ignatov2021learnedsmartphoneispmobile} provide smartphone RAW inputs and DSLR sRGB targets for RAW-to-sRGB translation, typically involving higher-quality devices and a wider variety of scenes. Despite these advances, existing RAW-to-sRGB datasets still face limitations such as cross-device misalignment and limited dynamic range. In this work, to advance realistic ISP research, we present LC-ISP, a spatially aligned live-and-capture stream ISP dataset that covers a broader and more diverse range of scenarios, including daytime and nighttime scenes, indoor and outdoor environments, and high dynamic range conditions.

\textbf{Learned image signal processing}
The image signal processing (ISP) pipeline converts RAW sensor data into sRGB images suitable for display or downstream vision tasks. Traditional ISP pipelines have been extensively studied and are typically composed of multiple stages, including demosaicking, denoising, high dynamic range imaging, white balancing, color correction and tone mapping \cite{demosaic,global_local_color,nonlocal,edge_color_constancy}. Numerous algorithms have been proposed to improve the performance of each individual stage, leading to continuous advances in image quality and efficiency. With the development of deep learning, deep networks have been exploited to tackle challenging ISP tasks such as denoising, color constancy and tone mapping \cite{chen2022simplebaselinesimagerestoration,color_xiao,lou2025learning,sgn,wang2020practicaldeeprawimage,Zhang_2018,pft,atd,dpe,hdrnet,contrast_enhancer,3dlut,Zhang_2017}. In addition, several recent works investigate how multiple restoration operations can be learned jointly as components of a single, unified framework \cite{michael_jdd,zhou2018deepresidualnetworkjoint,qian2023rethinkinglearningbaseddemosaicingdenoising}.

More recently, a number of works have attempted to replace the traditional ISP pipeline with a single deep model that performs end-to-end RAW-to-sRGB mapping. DeepISP \cite{Schwartz_2019} produces high-quality sRGB images from RAW inputs under challenging conditions such as low light. MWISPNet \cite{ignatov2020aim2020challengelearned} employs wavelet transforms and residual blocks to better preserve fine-grained structures from the RAW signal, while AWNet \cite{dai2020awnetattentivewaveletnetwork} integrates a global context block to achieve more effective color adjustment. To address the misalignment between mobile RAW and DSLR sRGB images in previous ISP datasets, LiteISPNet \cite{zhang2021learningrawtosrgbmappingsinaccurately} introduces an explicit alignment mechanism and a mapping strategy to produce high-quality, well-aligned sRGB outputs. 
Beyond perceptual image enhancement, several studies further tailor ISP models to downstream vision tasks, such as object detection and semantic segmentation \cite{chen2025taskawareimagesignalprocessor, cui2025rawadapteradaptingpretrainedvisual, raod,wang2024adaptiveisplearningadaptiveimage}.
In this paper, we introduce a two-stage model, Dual-Stage ISP, which effectively leverages real-world LC-ISP data for supervision. By designing task-specific network architectures for restoration and enhancement, we significantly improve single-frame imaging quality, enabling more robust and powerful real-time ISP capabilities.
\section{Live-and-capture stream ISP dataset}
\begin{table*}[t]
  \centering
  \caption{Comparison of the proposed LC-ISP with existing ISP datasets.}
  \label{dataset-comparison}

  \small
  \setlength{\tabcolsep}{1mm}
  \begin{tabular}{@{}lccccccc@{}}
    \toprule
    Dataset & Images & Illumination & HDR Scenes & Aligned Pairs & Environment & Patch Size & Bit Depth \\
    \midrule
    ZRR \cite{ignatov2020replacingmobilecameraisp} & 48,043 & Day & \xmark & \xmark & Outdoor & 448$\times$448 & 10 \\
    MAI \cite{ignatov2021learnedsmartphoneispmobile} & 24,100 & Day & \xmark & \xmark & Outdoor & 224$\times$224 & 12 \\
    \textbf{LC-ISP (Ours)} & \textbf{53,892} & \textbf{Day \& Night} & \checkmark & \checkmark & \textbf{In/Outdoor} & \textbf{448$\times$448} & \textbf{12} \\
    \bottomrule
  \end{tabular}
\end{table*}
\begin{figure*}[t]
    \centering
    \includegraphics[width=\textwidth]{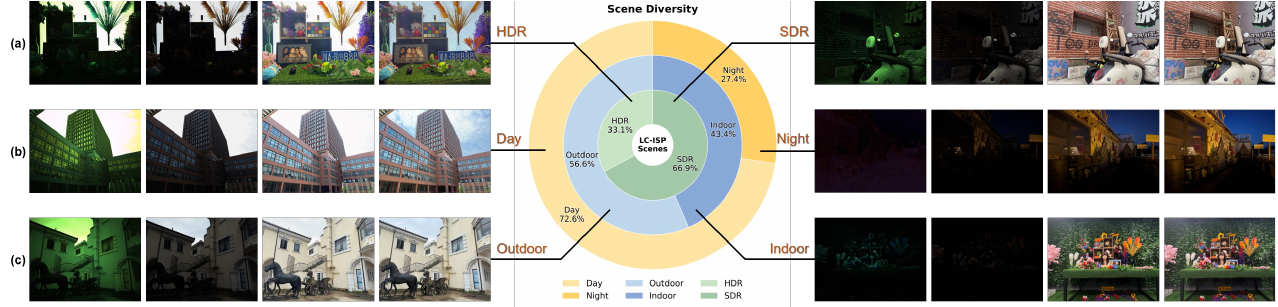}
    \vspace{-10pt}
    \caption{Examples from the LC-ISP Dataset. The figure shows the dataset's scene distribution and several groups of sample images, with each group arranged from left to right: single-frame RAW, multi-frame fused RGB, live sRGB, and capture sRGB. (a) HDR/SDR scenes with different dynamic ranges. (b) Day/Night scenes under varying illumination. (c) Outdoor/Indoor scenes in diverse environments.}
    \label{dataset}
    \vspace{-10pt}
\end{figure*}

\subsection{Dataset construction}
We captured a total of 1,000 distinct scenes using an iPhone 16 Pro under both daytime and nighttime conditions. The scenes were divided into training and test sets at a 9:1 ratio, after which each full-resolution capture (4032$\times$3024 pixels) was uniformly cropped into non-overlapping 448$\times$448 patches. This process yielded a total of 53,892 spatially aligned sample groups.
For each scene, we recorded two parallel ISP streams. The live stream provides a single RAW frame and its low-latency sRGB output. The capture stream aligns and fuses multiple RAW frames to produce an intermediate RGB image and the final capture sRGB ground truth. The intermediate RGB is obtained from Apple ProRAW, whose computational photography pipeline includes demosaicing, multi-frame denoising, HDR reconstruction, and white balance correction.
To obtain clean, high-SNR RGB and sRGB ground truth in the capture stream, we reduce ISO and proportionally increase exposure time while keeping the exposure value approximately constant. The ground-truth ISO ranges from 32 to 500, and the exposure time ranges from 1/2000 s to 1 s. All data were collected on a tripod under static conditions to reduce motion blur.
After capture, we manually inspected all potentially motion-affected regions under high zoom and discarded pairs with observable local motion, temporal shifts, or fusion artifacts. We also used a script to verify the consistency of white balance and overall exposure value between the live and capture streams. As illustrated in Figure~\ref{dataset}, LC-ISP spans a particularly broad variety of mobile photography scenarios, including indoor and outdoor scenes, low-light settings, and high-dynamic-range conditions. To the best of our knowledge, LC-ISP is the first large-scale dataset to explicitly encompass both live and capture streams for research on mobile image signal processing.

\begin{figure}[t]
    \centering
    \includegraphics[width=0.98\columnwidth]{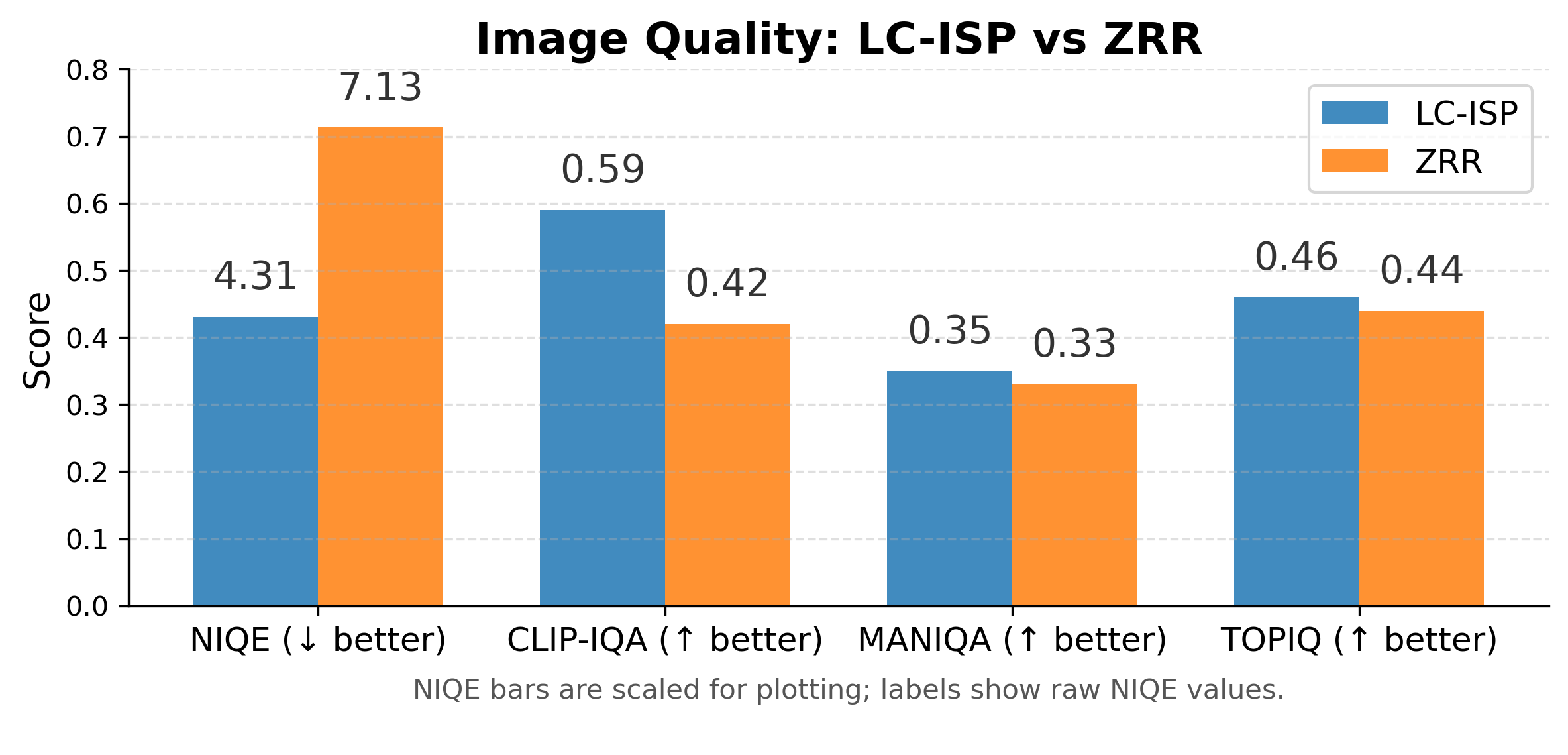}
    \vspace{-10pt}
\caption{Image-quality comparison on LC-ISP. The chart compares LC-ISP ground truth with ZRR \cite{ignatov2020replacingmobilecameraisp}, demonstrating the superior overall image quality of LC-ISP.}
    \label{stat}
\vspace{-10pt}
\end{figure}

\subsection{Comparison to existing datasets}
Table~\ref{dataset-comparison} compares LC-ISP with two representative ISP benchmarks: ZRR \cite{ignatov2020replacingmobilecameraisp} and MAI \cite{ignatov2021learnedsmartphoneispmobile}. LC-ISP not only contains more samples but also incorporates nighttime, indoor, and high-dynamic-range scenes, offering substantially broader diversity in illumination and environmental conditions.
Unlike ZRR and MAI, which rely on cross-device acquisition by pairing smartphone RAW inputs with DSLR outputs, LC-ISP records both live and capture streams from the same camera sensor, ensuring pixel-level spatial alignment for supervised learning. This same-device dual-pipeline design pairs a low-latency single-frame live stream with a high-quality multi-frame capture stream, preserving alignment while introducing a clear supervision-quality gap. Moreover, LC-ISP provides intermediate fused RGB outputs from the capture stream, enabling modular supervision and facilitating new research directions such as stage-wise ISP learning.
To evaluate the image quality of our ground-truth data, we conduct a quantitative no-reference IQA evaluation using NIQE \cite{niqe}, CLIP-IQA \cite{wang2022exploringclipassessinglook}, MANIQA \cite{yang2022maniqamultidimensionattentionnetwork}, and TOPIQ \cite{chen2023topiqtopdownapproachsemantics}. Since the MAI dataset is not publicly available, we compare against the ground truth from ZRR. These metrics compare reference-image quality across datasets without requiring spatially aligned image pairs. As shown in Figure~\ref{stat}, LC-ISP consistently outperforms ZRR in perceptual quality. Collectively, these characteristics establish LC-ISP as a comprehensive benchmark for advancing data-driven mobile ISP research under realistic conditions encountered in mobile image capture.

\begin{figure*}[t]
    \centering
    \includegraphics[width=\textwidth]{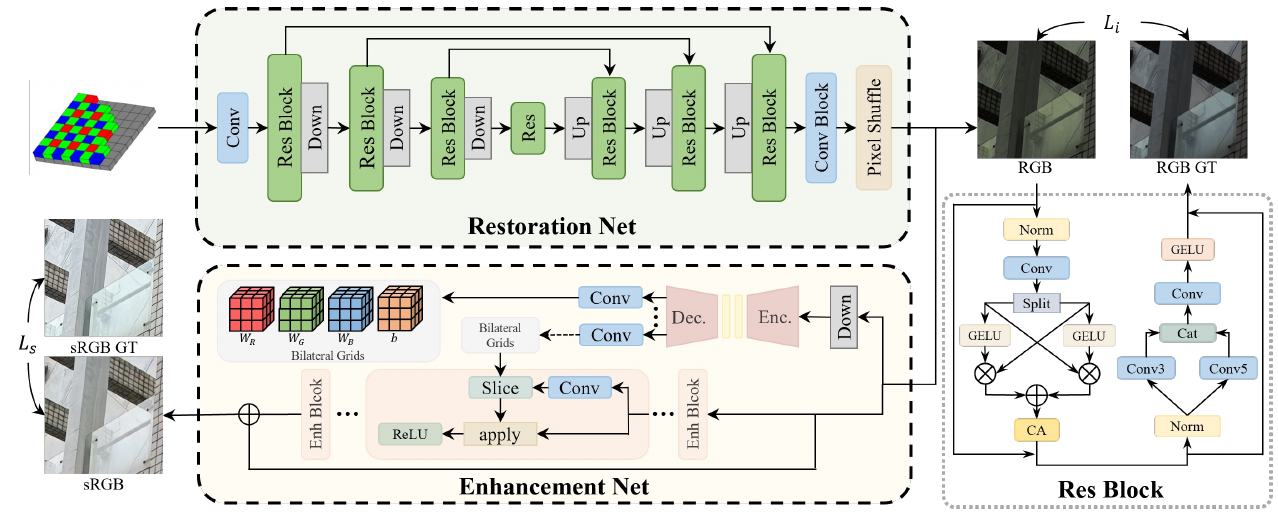}
    \vspace{-10pt}
    \caption{The overall architecture of our Dual-Stage ISP framework, consisting of a restoration network and an enhancement network.
Leveraging multi-frame fused RGB as prior information, the restoration network emulates the restored RGB of the capture stream. With a cleaner intermediate RGB, the pixel-adaptive enhancement block enables more flexible mapping, producing sRGB results that are perceptually closer to those of the capture stream.}
    \label{framework}
    \vspace{-10pt}
\end{figure*}

\section{Dual-Stage ISP}

\subsection{Overview}
Designing a single-frame ISP for mobile devices requires balancing image quality and computational efficiency under tight compute and memory budgets. Prior approaches often use monolithic end-to-end networks for RAW-to-sRGB mapping \cite{ignatov2020replacingmobilecameraisp,dai2020awnetattentivewaveletnetwork,zhang2021learningrawtosrgbmappingsinaccurately,ignatov2020aim2020challengelearned}, yet the limited dynamic range and signal-to-noise ratio of a RAW frame make such models ill-suited to capture the full ISP mapping at high fidelity. To obtain high-quality results from a single input while remaining practical on mobile hardware, we adopt a modular, task-decoupled design inspired by classical ISPs and incorporate multi-frame supervision into training: a compact restoration stage that approximates the information provided by multi-frame fusion, followed by an efficient enhancement stage that adaptively renders the final display-ready sRGB.
Concretely, we design \textbf{Dual-Stage ISP}, which decomposes the ISP into two jointly trained components. The restoration network $\mathcal{R}$ takes the packed RAW input $I$ and produces a clean intermediate $\tilde{R}$ that approximates the LC-ISP multi-frame fused RGB. The enhancement module $\mathcal{E}$ then applies spatially adaptive transforms to $\tilde{R}$ to generate the final sRGB output $\hat{S}$:
$$
\tilde{R}=\mathcal{R}(I), \qquad \hat{S}=\mathcal{E}(\tilde{R}).
$$
Here, $I$ denotes the packed RAW input, $\tilde{R}$ is the restored RGB intermediate, and $\hat{S}$ is the display-ready sRGB image. During training, $\mathcal{R}$ is supervised with multi-frame fused RGB and $\mathcal{E}$ with capture sRGB from LC-ISP; at test time, only a single RAW frame is required. This decomposition concentrates prior learning in training, narrows the single–to-multi gap through intermediate supervision, and ensures the model remains within the feasible computational limits for realistic mobile ISP deployment.

\subsection{Restoration net}
The restoration module \(\mathcal{R}\) is designed to emulate multi-frame fusion and generate a reliable intermediate representation. In our implementation, \(\mathcal{R}\) adopts an encoder–decoder topology inspired by U-Net \cite{ronneberger2015unetconvolutionalnetworksbiomedical}. It first uses a projection convolution to lift packed RAW inputs into an internal feature space, followed by downsampling encoder blocks, bottleneck blocks, and symmetric decoder blocks with skip connections and pixel-shuffle upsampling.

The core component of the restoration network is the Restoration Block (Res Block), which reconstructs high-SNR representations that approximate multi-frame fusion from a single RAW input (see Figure~\ref{framework}). Each block contains two cooperating units: a global reweighting unit and a local refinement unit. The global reweighting unit expands features into a higher-dimensional space and applies a nongating mechanism with channel attention to adaptively reweight channel responses. This simulates multi-frame feature distributions and suppresses sensor noise, producing cleaner structural representations. The local refinement unit then employs multi-scale depthwise convolutions to exploit local similarity priors, followed by a convolutional fusion layer to aggregate spatial information, refine textures, and remove residual noise. Overall, this design balances noise suppression and detail preservation in single-frame restoration, emulating the benefits of multi-frame fusion while maintaining computational efficiency during restoration.

\subsection{Enhancement net}
The final rendering from restored RGB to display-ready sRGB requires both global re-rendering and fine-grained, content-aware adjustments. Per-pixel transformations are particularly effective because they enable spatially adaptive color and tone corrections while remaining computationally cheap. Inspired by BPAM \cite{lou2025learning}, we design a multi-stage pixel-wise enhancement network that performs hierarchical, nonlinear adjustments for precise color and tone rendering. In our design, the network first produces compact, multi-scale bilateral grids via an encoder–decoder operating on a low-resolution version of the image; a parallel branch extracts full-resolution context maps that capture scene-dependent cues. The context maps are then used to slice the grids and generate pixel-wise transforms that are applied directly in the image domain across several successive stages. By stacking lightweight pixel-wise transforms rather than computing large latent coefficients, the enhancement network achieves strong nonlinear mapping capacity with much lower computation, yielding flexible, spatially-varying rendering suitable for practical deployment in live imaging pipelines on mobile devices.

\subsection{Loss function}
Our model is trained with two supervision branches applied in parallel: an intermediate supervision on the restored output and a final supervision on the rendered sRGB output.
For the final output, we define the composite loss \(L_{s}\), consisting of pixel, perceptual, and structural terms:
\[
L_{s}=L_{\ell_1}+\lambda_{vgg}L_{\mathrm{vgg}}+\lambda_{\mathrm{ssim}}L_{\mathrm{ssim}},
\]
where \(L_{\ell_1}=\lVert \hat{S}-S\rVert_1\) is the pixel loss,
\(L_{\mathrm{vgg}}=\lVert \Phi(\hat{S})-\Phi(S)\rVert_1\) is the perceptual loss computed on VGG-19 \cite{simonyan2014very} feature maps \(\Phi(\cdot)\),
and \(L_{\mathrm{ssim}}=1-\mathrm{MS\_SSIM}(\hat{S},S)\) is the loss term based on structural similarity.

For the intermediate restoration output $\tilde{R}$, we use a restoration loss $L_i$ that combines a pixel-wise L1 term with Sobel-based edge-aware regularization:
$$
L_{i} = L_{i}^{\ell_1} + \lambda_{\mathrm{edge}}L_{\mathrm{edge}} + \lambda^{(i)}_{vgg} L_{vgg},
$$
where the Sobel-based edge term is defined as
$$
L_{\mathrm{edge}} = \lVert \mathcal{S}(\tilde{R}) - \mathcal{S}(R_{\mathrm{ref}}) \rVert_1.
$$
$\mathcal{S}(\cdot)$ denotes the Sobel operator that computes horizontal and vertical image gradients. We apply $\mathcal{S}$ to the luminance channel and use the L1 difference between gradient maps to encourage matching the reference image in both high-frequency structure and low-frequency content.

The total training objective explicitly integrates both intermediate supervision and final supervision:
$$
L_{\mathrm{total}} = L_{s} + \lambda_{i}L_{i},
$$
where $\lambda_{i}$ balances the influence of the intermediate guidance.
This joint objective guides the restoration module toward clean, high-SNR predictions while ensuring perceptually faithful and structurally consistent sRGB outputs.
\begin{table*}[t]
  \centering
  \caption{Quantitative results for single-frame ISP models of different scales
  on the LC-ISP dataset. GFLOPs are measured using $448 \times 448$ RAW inputs.
  Best results are in \textbf{bold}.}
  \label{tab:quantitative}

  \begingroup
  \small
  \setlength{\tabcolsep}{8pt}

  \begin{tabular}{@{}llccccc@{}}
    \toprule
    Category & Model
    & PSNR $\uparrow$
    & SSIM $\uparrow$
    & \shortstack{MS-SSIM $\uparrow$}
    & LPIPS $\downarrow$
    & GFLOPs \\
    \midrule

    \multirow{7}{*}{Base}
      & PyNet \cite{ignatov2020replacingmobilecameraisp}
      & 23.28 & 0.7948 & 0.9091 & 0.2531 & 685.16 \\
      & AWNet \cite{dai2020awnetattentivewaveletnetwork}
      & 24.43 & 0.7970 & 0.9089 & 0.2561 & 131.96 \\
      & MWISPNet \cite{ignatov2020aim2020challengelearned}
      & 24.72 & 0.8020 & 0.9120 & 0.2495 & 176.80 \\
      & LiteISPNet \cite{zhang2021learningrawtosrgbmappingsinaccurately}
      & 24.73 & 0.8028 & 0.9122 & 0.2477 & 143.89 \\
      & SYENet \cite{gou2023syenetsimpleeffectivenetwork}
      & 20.00 & 0.7178 & 0.8561 & 0.3773 & 10.20 \\
      & MobileIE \cite{yan2025mobileie}
      & 19.27 & 0.7111 & 0.8539 & 0.3937 & 4.96 \\
      & \textbf{Ours}
      & \textbf{25.36}
      & \textbf{0.8049}
      & \textbf{0.9138}
      & \textbf{0.2375}
      & \textbf{42.58} \\

    \midrule

    \multirow{5}{*}{Medium}
      & PyNet \cite{ignatov2020replacingmobilecameraisp}
      & 20.52 & 0.7444 & 0.8728 & 0.3641 & 10.84 \\
      & AWNet \cite{dai2020awnetattentivewaveletnetwork}
      & 21.94 & 0.7489 & 0.8778 & 0.3195 & 10.31 \\
      & MWISPNet \cite{ignatov2020aim2020challengelearned}
      & 23.37 & 0.7810 & 0.8989 & 0.2959 & 11.20 \\
      & LiteISPNet \cite{zhang2021learningrawtosrgbmappingsinaccurately}
      & 23.38 & 0.7842 & 0.8997 & 0.2910 & 10.63 \\
      & \textbf{Ours}
      & \textbf{24.66}
      & \textbf{0.7948}
      & \textbf{0.9076}
      & \textbf{0.2458}
      & \textbf{10.98} \\

    \midrule

    \multirow{5}{*}{Lite}
      & PyNet \cite{ignatov2020replacingmobilecameraisp}
      & 20.09 & 0.7239 & 0.8600 & 0.4062 & 2.75 \\
      & AWNet \cite{dai2020awnetattentivewaveletnetwork}
      & 19.79 & 0.7285 & 0.8644 & 0.3493 & 4.46 \\
      & MWISPNet \cite{ignatov2020aim2020challengelearned}
      & 21.37 & 0.7388 & 0.8745 & 0.3624 & 2.30 \\
      & LiteISPNet \cite{zhang2021learningrawtosrgbmappingsinaccurately}
      & 21.31 & 0.7350 & 0.8698 & 0.3858 & 2.27 \\
      & \textbf{Ours}
      & \textbf{23.36}
      & \textbf{0.7748}
      & \textbf{0.8934}
      & \textbf{0.2850}
      & \textbf{2.50} \\

    \bottomrule
  \end{tabular}

  \endgroup

\end{table*}
\begin{figure*}[t]
    \centering
    \includegraphics[width=0.6150\textwidth]{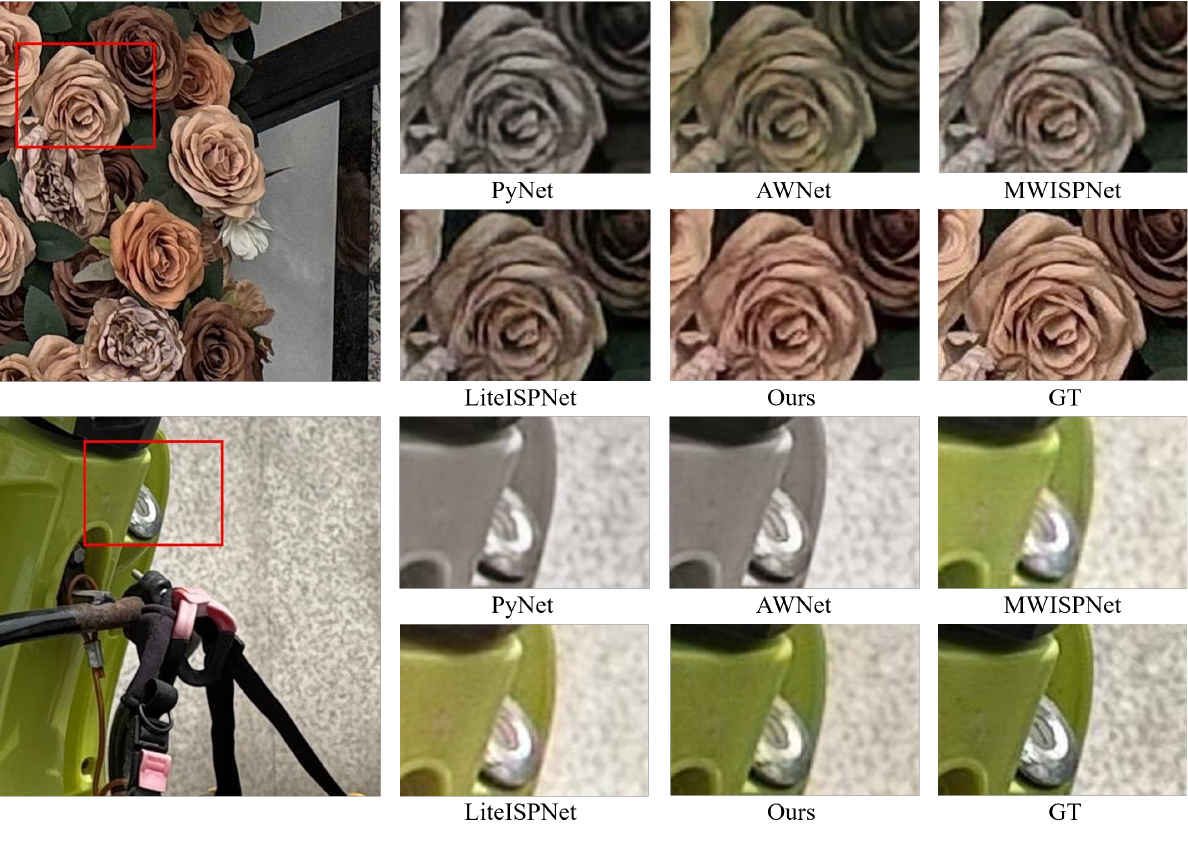}
\vspace{-8pt}
    \caption{From top to bottom: results for medium, and lite models. Our approach preserves more details, maintains balanced colors, and suppresses noise more effectively. Please zoom in for better view.}
    \label{result}
    \vspace{-0.5pt}
\end{figure*}

\section{Experiments}

\subsection{Experimental setup and implementation}
\paragraph{Dataset.}
We conduct experiments on LC-ISP using 48.5K training pairs and 5.3K test pairs. The input to the model is a single-frame RAW image, while the ground truth corresponds to the capture sRGB output.
\paragraph{Comparison methods.}
To demonstrate the effectiveness of our framework, we compare it with several state-of-the-art single-frame ISP methods, including PyNet, AWNet, MWISPNet, LiteISPNet, SYENet, MobileIE \cite{ignatov2020replacingmobilecameraisp,dai2020awnetattentivewaveletnetwork,ignatov2020aim2020challengelearned,zhang2021learningrawtosrgbmappingsinaccurately,gou2023syenetsimpleeffectivenetwork,yan2025mobileie}. We evaluate the reconstruction results with standard full-reference image quality metrics: PSNR, SSIM, MS-SSIM, and LPIPS \cite{zhang2018unreasonableeffectivenessdeepfeatures,msssim,ssim}.
\paragraph{Implementation details.}
All experiments are implemented in PyTorch and trained on four NVIDIA GeForce RTX 4090 GPUs for a total of 150 epochs. We use the Adam optimizer with a learning rate that is linearly warmed up from $1\times10^{-6}$ to $1\times10^{-4}$ over the first 5 epochs, and then decayed following a cosine schedule.
The batch size is set to 24. Data augmentation includes random horizontal and vertical flips as well as rotations. For fair comparison, all models are trained under identical settings.

\subsection{Experimental results}
\label{exp}
\paragraph{Quantitative comparison.}
We first compare our method with the original versions of several state-of-the-art single-frame ISPs to establish baseline performance. To evaluate mobile applicability, we further construct medium and lite variants of each model by reducing block counts or channel widths, ensuring comparable FLOPs across methods. Quantitative results are summarized in Table \ref{tab:quantitative}.

As shown in Table~\ref{tab:quantitative}, our method consistently outperforms existing single-frame ISPs in their original configurations, achieving higher performance with substantially fewer computations than strong baselines such as MWISPNet \cite{ignatov2020aim2020challengelearned} and LiteISPNet \cite{zhang2021learningrawtosrgbmappingsinaccurately}. Although SYENet \cite{gou2023syenetsimpleeffectivenetwork} and MobileIE \cite{yan2025mobileie} have the lowest computational costs, their reconstruction quality is significantly inferior. Among all methods, our two-stage framework transfers multi-frame priors to the single-frame live process, improving fidelity and stability without extra runtime overhead.
When models are scaled down for mobile constraints, our advantage becomes more evident. Under medium and lightweight budgets, competing architectures suffer larger performance drops, whereas our model maintains stronger fidelity and robustness. Overall, the proposed framework achieves a favorable balance between efficiency and image quality, performing well at full scale while preserving its advantages as model size decreases, making it suitable for power-constrained mobile scenarios.

\paragraph{Qualitative comparison.}
Figure~\ref{result} presents visual comparisons across medium and lite models. Our method better preserves fine details, maintains accurate color reproduction, and suppresses noise. At the medium scale, PyNet and AWNet show noticeable color desaturation, while MWISPNet and LiteISPNet introduce visible noise and texture distortions. Our results consistently maintain higher visual fidelity.
The gap becomes more pronounced in the lite regime, where competing methods suffer from color fading, residual noise, and detail loss, whereas our model retains clearer textures and more stable colors. These results show that our two-stage design degrades gracefully with reduced capacity, making it well suited for mobile live deployment under tight computational budgets. Additional visual comparisons are provided in the supplementary material.

\subsection{Dataset diversity and scalability analysis}
To further evaluate the representative capacity and diversity of the proposed LC-ISP dataset, we conduct a comparative analysis against the ZRR dataset \cite{ignatov2020replacingmobilecameraisp}. Using the MWISPNet \cite{ignatov2020aim2020challengelearned} as our benchmark architecture, we train the model on both datasets across varying data scales: 5\%, 10\%, 25\%, 50\%, and 100\%. To maintain the integrity of the comparison and mitigate selection bias, training samples for each subset were selected at random, with a uniform sampling probability for each training example.

\begin{figure}[t]
    \centering
    \includegraphics[width=0.98\columnwidth]{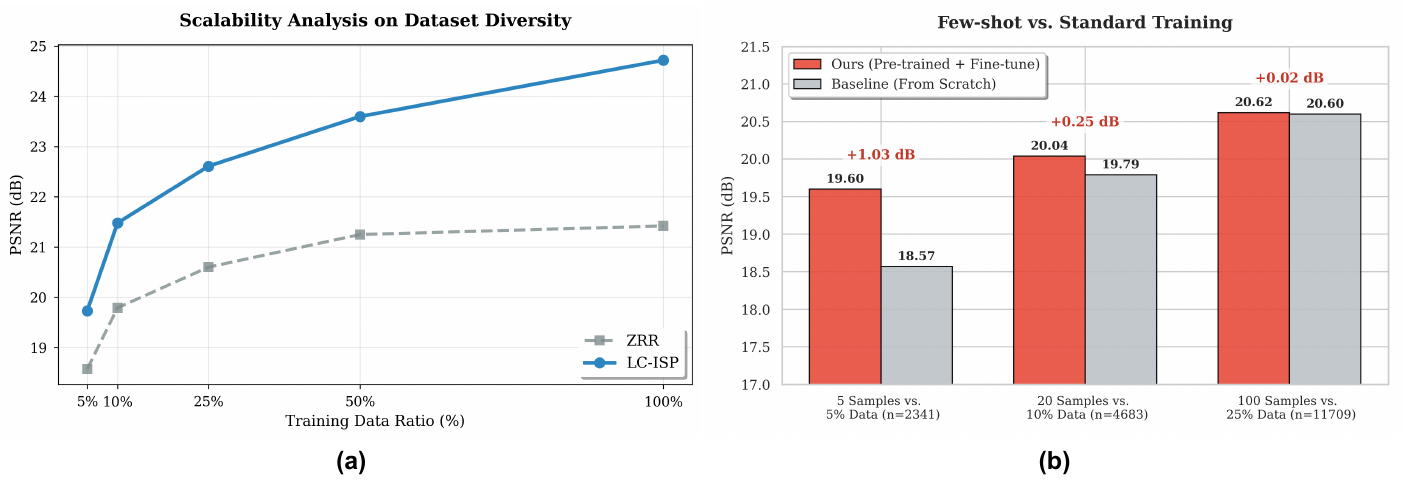}
\vspace{-10pt}
    \caption{(a) Performance across data ratios on LC-ISP and ZRR. (b) Few-shot adaptation against training from scratch.}
    \label{diversity}
\vspace{-8pt}
\end{figure}

As illustrated in the performance-to-data-ratio curves in Figure~\ref{diversity} (a), a distinct performance gap emerges between the two datasets. While the MWISPNet trained on the ZRR dataset exhibits an early performance plateau, suggesting a degree of informational redundancy, our LC-ISP dataset facilitates consistent and significant PSNR/SSIM improvements as the data volume scales. This sustained upward trajectory underscores the superior information density of LC-ISP. It implies that our dataset encapsulates a more diverse manifold of lighting conditions and intricate scene structures, which effectively prevents the model from over-fitting to limited patterns and encourages better generalization.

\subsection{Cross-dataset few-shot adaptation}
To evaluate the transferability of representations learned from LC-ISP, we conduct a cross-dataset few-shot adaptation study. MWISPNet \cite{ignatov2020aim2020challengelearned} is first pre-trained on LC-ISP and subsequently fine-tuned on ZRR \cite{ignatov2020replacingmobilecameraisp} using a limited number of target-domain samples. The fine-tuning stage is limited to 10 epochs, with all other optimization settings kept unchanged from the original training setup.

As shown in Figure~\ref{diversity} (b), the pre-trained model adapts efficiently to the new dataset and achieves strong restoration performance under sparse supervision. For comparison, models trained from scratch are optimized on proportionally sampled subsets of the ZRR training set. Despite being fine-tuned with substantially fewer samples and for a short training duration, the LC-ISP pre-trained model converges significantly faster and consistently achieves superior performance across all few-shot settings. This suggests that our dataset facilitates the extraction of universal image priors rather than dataset-specific artifacts. These results confirm that LC-ISP enables learning transferable low-level representations that generalize beyond the characteristics of a specific camera processing pipeline.

\subsection{Ablation study}
We conduct ablation studies on LC-ISP under identical training settings, first evaluating our two-stage framework and then testing its generalization across different choices of network backbone architecture.

\begin{figure}[t]
    \centering
    \includegraphics[width=0.96\columnwidth]{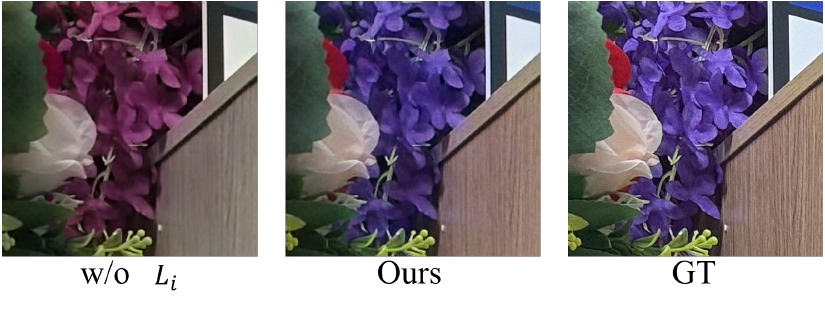}
\vspace{-10pt}
    \caption{Visual comparison of ablation results.
The proposed intermediate supervision helps preserve fine details and improve color fidelity, enhancing overall image quality.}
    \label{fig:ablation}
\vspace{-8pt}
\end{figure}

In Section~\ref{exp}, we have demonstrated that our framework outperforms existing state-of-the-art methods, including single-stage end-to-end ISPs. Here, we further analyze the proposed two-stage design by evaluating the role of intermediate supervision. Specifically, we conduct experiments on our lite model by removing the intermediate supervision term. As shown in Table~\ref{abl}, even without intermediate supervision, our two-stage framework already surpasses the best-performing single-stage ISP. With intermediate supervision, which leverages additional multi-frame information during training, the model achieves further gains.
The visualization in Figure~\ref{fig:ablation} shows that intermediate supervision helps preserve fine details and produce more accurate, visually pleasing colors.
\begin{table}[t]
\centering
\caption{Ablation study of the proposed two-stage framework. Intermediate supervision improves performance by leveraging multi-frame information, while the last two rows show that the design also benefits other architectures.}
\label{abl}
\setlength{\tabcolsep}{2.5pt}
\small

\begin{tabular}{l|cccc}
\toprule
Method & PSNR $\uparrow$ & SSIM $\uparrow$ & MS-SSIM $\uparrow$ & LPIPS $\downarrow$ \\
\midrule
MWISPNet & 21.37 & 0.7388 & 0.8745 & 0.3624 \\
Ours w/o $L_i$ & 22.78 & 0.7693 & 0.8894 & 0.2980 \\
Ours (Full) & \textbf{23.36} & \textbf{0.7748} & \textbf{0.8934} & \textbf{0.2850} \\
\midrule
NAF+CSR w/o $L_i$ &21.91 & 0.7556&0.8748 & 0.3121 \\
NAF+CSR  & 22.56& 0.7648& 0.8869&  0.3037\\
\bottomrule
\end{tabular}
\end{table}

To further assess the robustness and generality of our two-stage framework, we extend it to alternative backbone architectures. In particular, we adopt NAFNet~\cite{chen2022simplebaselinesimagerestoration} as the restoration network and CSRNet~\cite{he2020conditionalsequentialmodulationefficient} as the enhancement network. As shown in Table~\ref{abl}, this configuration consistently benefits from the two-stage structure, confirming the framework’s general applicability. However, its performance remains inferior to our default architecture, highlighting the effectiveness of our task-aware design.

\section{Conclusion}
In this work, we introduce LC-ISP, a large-scale live-and-capture stream benchmark for data-driven mobile ISP. By leveraging the intrinsic gap between live and capture streams, LC-ISP provides spatially aligned, realistic supervision without cross-device inconsistencies, while covering diverse scenes, illumination conditions, and intermediate fused outputs. Experiments with a compact two-stage baseline show that capture-stream intermediates effectively improve single-frame RAW-to-sRGB mapping, producing sharper details, lower noise, and more faithful colors under practical computational budgets. We hope LC-ISP will support future research into the design of lightweight ISP architectures and the development of learning paradigms for next-generation mobile imaging systems.

\section*{Acknowledgement}
This work was supported by the National Natural Science
Foundation of China (No. 62476051) and Sichuan Natural
Science Foundation (No. 2024NSFTD0041).
\newpage
{\small
\bibliographystyle{ieeenat_fullname}
\bibliography{references}

@misc{chen2022simplebaselinesimagerestoration,
      title={Simple Baselines for Image Restoration}, 
      author={Liangyu Chen and Xiaojie Chu and Xiangyu Zhang and Jian Sun},
      year={2022},
      eprint={2204.04676},
      archivePrefix={arXiv},
      primaryClass={cs.CV},
      url={https://arxiv.org/abs/2204.04676}, 
}

@INPROCEEDINGS{color_xiao,
  author={Xiao, Jin and Gu, Shuhang and Zhang, Lei},
  booktitle={2020 IEEE/CVF Conference on Computer Vision and Pattern Recognition (CVPR)}, 
  title={Multi-Domain Learning for Accurate and Few-Shot Color Constancy}, 
  year={2020},
  volume={},
  number={},
  pages={3255-3264},
  doi={10.1109/CVPR42600.2020.00332}}

@inproceedings{lou2025learning,
  title={Learning Pixel-adaptive Multi-layer Perceptrons for Real-time Image Enhancement},
  author={Lou, Junyu and Zhao, Xiaorui and Shi, Kexuan and Gu, Shuhang},
  booktitle={Proceedings of the IEEE/CVF International Conference on Computer Vision},
  pages={14095--14105},
  year={2025}
}

@misc{chen2018learningdark,
      title={Learning to See in the Dark}, 
      author={Chen Chen and Qifeng Chen and Jia Xu and Vladlen Koltun},
      year={2018},
      eprint={1805.01934},
      archivePrefix={arXiv},
      primaryClass={cs.CV},
      url={https://arxiv.org/abs/1805.01934}, 
}

@misc{dnd,
      title={Benchmarking Denoising Algorithms with Real Photographs}, 
      author={Tobias Plötz and Stefan Roth},
      year={2017},
      eprint={1707.01313},
      archivePrefix={arXiv},
      primaryClass={cs.CV},
      url={https://arxiv.org/abs/1707.01313}, 
}

@misc{wang2020practicaldeeprawimage,
      title={Practical Deep Raw Image Denoising on Mobile Devices}, 
      author={Yuzhi Wang and Haibin Huang and Qin Xu and Jiaming Liu and Yiqun Liu and Jue Wang},
      year={2020},
      eprint={2010.06935},
      archivePrefix={arXiv},
      primaryClass={eess.IV},
      url={https://arxiv.org/abs/2010.06935}, 
}

@INPROCEEDINGS{sgn,
  author={Gu, Shuhang and Li, Yawei and Van Gool, Luc and Timofte, Radu},
  booktitle={2019 IEEE/CVF International Conference on Computer Vision (ICCV)}, 
  title={Self-Guided Network for Fast Image Denoising}, 
  year={2019},
  volume={},
  number={},
  pages={2511-2520},
  doi={10.1109/ICCV.2019.00260}}

@INPROCEEDINGS{nonlocal,
  author={Buades, A. and Coll, B. and Morel, J.-M.},
  booktitle={2005 IEEE Computer Society Conference on Computer Vision and Pattern Recognition (CVPR'05)}, 
  title={A non-local algorithm for image denoising}, 
  year={2005},
  volume={2},
  number={},
  pages={60-65 vol. 2},
  doi={10.1109/CVPR.2005.38}}

@ARTICLE{demosaic,
  author={Hirakawa, K. and Parks, T.W.},
  journal={IEEE Transactions on Image Processing}, 
  title={Adaptive homogeneity-directed demosaicing algorithm}, 
  year={2005},
  volume={14},
  number={3},
  pages={360-369},
  doi={10.1109/TIP.2004.838691}}

@article{global_local_color,
author = {Rizzi, Alessandro and Gatta, Carlo and Marini, Daniele},
title = {A new algorithm for unsupervised global and local color correction},
year = {2003},
issue_date = {July 2003},
publisher = {Elsevier Science Inc.},
address = {USA},
volume = {24},
number = {11},
issn = {0167-8655},
url = {https://doi.org/10.1016/S0167-8655(02)00323-9},
doi = {10.1016/S0167-8655(02)00323-9},
journal = {Pattern Recogn. Lett.},
month = jul,
pages = {1663–1677},
numpages = {15}
}

@ARTICLE{edge_color_constancy,
  author={van de Weijer, Joost and Gevers, Theo and Gijsenij, Arjan},
  journal={IEEE Transactions on Image Processing}, 
  title={Edge-Based Color Constancy}, 
  year={2007},
  volume={16},
  number={9},
  pages={2207-2214},
  doi={10.1109/TIP.2007.901808}}

@article{Schwartz_2019,
   title={DeepISP: Toward Learning an End-to-End Image Processing Pipeline},
   volume={28},
   ISSN={1941-0042},
   url={http://dx.doi.org/10.1109/TIP.2018.2872858},
   DOI={10.1109/tip.2018.2872858},
   number={2},
   journal={IEEE Transactions on Image Processing},
   publisher={Institute of Electrical and Electronics Engineers (IEEE)},
   author={Schwartz, Eli and Giryes, Raja and Bronstein, Alex M.},
   year={2019},
   month=feb, pages={912–923} }

@misc{ignatov2020aim2020challengelearned,
      title={AIM 2020 Challenge on Learned Image Signal Processing Pipeline}, 
      author={Andrey Ignatov and Radu Timofte and Zhilu Zhang and Ming Liu and Haolin Wang and Wangmeng Zuo and Jiawei Zhang and Ruimao Zhang and Zhanglin Peng and Sijie Ren and Linhui Dai and Xiaohong Liu and Chengqi Li and Jun Chen and Yuichi Ito and Bhavya Vasudeva and Puneesh Deora and Umapada Pal and Zhenyu Guo and Yu Zhu and Tian Liang and Chenghua Li and Cong Leng and Zhihong Pan and Baopu Li and Byung-Hoon Kim and Joonyoung Song and Jong Chul Ye and JaeHyun Baek and Magauiya Zhussip and Yeskendir Koishekenov and Hwechul Cho Ye and Xin Liu and Xueying Hu and Jun Jiang and Jinwei Gu and Kai Li and Pengliang Tan and Bingxin Hou},
      year={2020},
      eprint={2011.04994},
      archivePrefix={arXiv},
      primaryClass={cs.CV},
      url={https://arxiv.org/abs/2011.04994}, 
}

@misc{ignatov2020replacingmobilecameraisp,
      title={Replacing Mobile Camera ISP with a Single Deep Learning Model}, 
      author={Andrey Ignatov and Luc Van Gool and Radu Timofte},
      year={2020},
      eprint={2002.05509},
      archivePrefix={arXiv},
      primaryClass={cs.CV},
      url={https://arxiv.org/abs/2002.05509}, 
}

@misc{dai2020awnetattentivewaveletnetwork,
      title={AWNet: Attentive Wavelet Network for Image ISP}, 
      author={Linhui Dai and Xiaohong Liu and Chengqi Li and Jun Chen},
      year={2020},
      eprint={2008.09228},
      archivePrefix={arXiv},
      primaryClass={eess.IV},
      url={https://arxiv.org/abs/2008.09228}, 
}

@misc{zhang2021learningrawtosrgbmappingsinaccurately,
      title={Learning RAW-to-sRGB Mappings with Inaccurately Aligned Supervision}, 
      author={Zhilu Zhang and Haolin Wang and Ming Liu and Ruohao Wang and Jiawei Zhang and Wangmeng Zuo},
      year={2021},
      eprint={2108.08119},
      archivePrefix={arXiv},
      primaryClass={cs.CV},
      url={https://arxiv.org/abs/2108.08119}, 
}

@misc{ronneberger2015unetconvolutionalnetworksbiomedical,
      title={U-Net: Convolutional Networks for Biomedical Image Segmentation}, 
      author={Olaf Ronneberger and Philipp Fischer and Thomas Brox},
      year={2015},
      eprint={1505.04597},
      archivePrefix={arXiv},
      primaryClass={cs.CV},
      url={https://arxiv.org/abs/1505.04597}, 
}

@misc{gou2023syenetsimpleeffectivenetwork,
      title={SYENet: A Simple Yet Effective Network for Multiple Low-Level Vision Tasks with Real-time Performance on Mobile Device}, 
      author={Weiran Gou and Ziyao Yi and Yan Xiang and Shaoqing Li and Zibin Liu and Dehui Kong and Ke Xu},
      year={2023},
      eprint={2308.08137},
      archivePrefix={arXiv},
      primaryClass={cs.CV},
      url={https://arxiv.org/abs/2308.08137}, 
}

@ARTICLE{ssim,
  author={Zhou Wang and Bovik, A.C. and Sheikh, H.R. and Simoncelli, E.P.},
  journal={IEEE Transactions on Image Processing}, 
  title={Image quality assessment: from error visibility to structural similarity}, 
  year={2004},
  volume={13},
  number={4},
  pages={600-612},
  doi={10.1109/TIP.2003.819861}}

@misc{zhang2018unreasonableeffectivenessdeepfeatures,
      title={The Unreasonable Effectiveness of Deep Features as a Perceptual Metric}, 
      author={Richard Zhang and Phillip Isola and Alexei A. Efros and Eli Shechtman and Oliver Wang},
      year={2018},
      eprint={1801.03924},
      archivePrefix={arXiv},
      primaryClass={cs.CV},
      url={https://arxiv.org/abs/1801.03924}, 
}

@INPROCEEDINGS{msssim,
  author={Wang, Z. and Simoncelli, E.P. and Bovik, A.C.},
  booktitle={The Thrity-Seventh Asilomar Conference on Signals, Systems \& Computers, 2003}, 
  title={Multiscale structural similarity for image quality assessment}, 
  year={2003},
  volume={2},
  number={},
  pages={1398-1402 Vol.2},
  doi={10.1109/ACSSC.2003.1292216}}

@article{Zhang_2017,
   title={Beyond a Gaussian Denoiser: Residual Learning of Deep CNN for Image Denoising},
   volume={26},
   ISSN={1941-0042},
   url={http://dx.doi.org/10.1109/TIP.2017.2662206},
   DOI={10.1109/tip.2017.2662206},
   number={7},
   journal={IEEE Transactions on Image Processing},
   publisher={Institute of Electrical and Electronics Engineers (IEEE)},
   author={Zhang, Kai and Zuo, Wangmeng and Chen, Yunjin and Meng, Deyu and Zhang, Lei},
   year={2017},
   month=jul, pages={3142–3155} }

@article{kalantari_hdr,
author = {Kalantari, Nima Khademi and Ramamoorthi, Ravi},
title = {Deep high dynamic range imaging of dynamic scenes},
year = {2017},
issue_date = {August 2017},
publisher = {Association for Computing Machinery},
address = {New York, NY, USA},
volume = {36},
number = {4},
issn = {0730-0301},
url = {https://doi.org/10.1145/3072959.3073609},
doi = {10.1145/3072959.3073609},
journal = {ACM Trans. Graph.},
month = jul,
articleno = {144},
numpages = {12}
}

@INPROCEEDINGS{dped,
  author={Ignatov, Andrey and Kobyshev, Nikolay and Timofte, Radu and Vanhoey, Kenneth},
  booktitle={2017 IEEE International Conference on Computer Vision (ICCV)}, 
  title={DSLR-Quality Photos on Mobile Devices with Deep Convolutional Networks}, 
  year={2017},
  volume={},
  number={},
  pages={3297-3305},
  doi={10.1109/ICCV.2017.355}}

@INPROCEEDINGS{mit5k,
  author={Bychkovsky, Vladimir and Paris, Sylvain and Chan, Eric and Durand, Fredo},
  booktitle={CVPR 2011}, 
  title={Learning photographic global tonal adjustment with a database of input / output image pairs}, 
  year={2011},
  volume={},
  number={},
  pages={97-104},
  doi={10.1109/CVPR.2011.5995332}}

@article{michael_jdd,
author = {Gharbi, Micha\"{e}l and Chaurasia, Gaurav and Paris, Sylvain and Durand, Fr\'{e}do},
title = {Deep joint demosaicking and denoising},
year = {2016},
issue_date = {November 2016},
publisher = {Association for Computing Machinery},
address = {New York, NY, USA},
volume = {35},
number = {6},
issn = {0730-0301},
url = {https://doi.org/10.1145/2980179.2982399},
doi = {10.1145/2980179.2982399},
journal = {ACM Trans. Graph.},
month = dec,
articleno = {191},
numpages = {12}
}

@misc{zhou2018deepresidualnetworkjoint,
      title={Deep Residual Network for Joint Demosaicing and Super-Resolution}, 
      author={Ruofan Zhou and Radhakrishna Achanta and Sabine Süsstrunk},
      year={2018},
      eprint={1802.06573},
      archivePrefix={arXiv},
      primaryClass={cs.CV},
      url={https://arxiv.org/abs/1802.06573}, 
}

@misc{qian2023rethinkinglearningbaseddemosaicingdenoising,
      title={Rethinking Learning-based Demosaicing, Denoising, and Super-Resolution Pipeline}, 
      author={Guocheng Qian and Yuanhao Wang and Jinjin Gu and Chao Dong and Wolfgang Heidrich and Bernard Ghanem and Jimmy S. Ren},
      year={2023},
      eprint={1905.02538},
      archivePrefix={arXiv},
      primaryClass={eess.IV},
      url={https://arxiv.org/abs/1905.02538}, 
}

@article{Zhang_2018,
   title={FFDNet: Toward a Fast and Flexible Solution for CNN-Based Image Denoising},
   volume={27},
   ISSN={1941-0042},
   url={http://dx.doi.org/10.1109/TIP.2018.2839891},
   DOI={10.1109/tip.2018.2839891},
   number={9},
   journal={IEEE Transactions on Image Processing},
   publisher={Institute of Electrical and Electronics Engineers (IEEE)},
   author={Zhang, Kai and Zuo, Wangmeng and Zhang, Lei},
   year={2018},
   month=sep, pages={4608–4622} }

@INPROCEEDINGS{dpe,
  author={Chen, Yu-Sheng and Wang, Yu-Ching and Kao, Man-Hsin and Chuang, Yung-Yu},
  booktitle={2018 IEEE/CVF Conference on Computer Vision and Pattern Recognition}, 
  title={Deep Photo Enhancer: Unpaired Learning for Image Enhancement from Photographs with GANs}, 
  year={2018},
  volume={},
  number={},
  pages={6306-6314},
  doi={10.1109/CVPR.2018.00660}}

@article{hdrnet,
   title={Deep bilateral learning for real-time image enhancement},
   volume={36},
   ISSN={1557-7368},
   url={http://dx.doi.org/10.1145/3072959.3073592},
   DOI={10.1145/3072959.3073592},
   number={4},
   journal={ACM Transactions on Graphics},
   publisher={Association for Computing Machinery (ACM)},
   author={Gharbi, Michaël and Chen, Jiawen and Barron, Jonathan T. and Hasinoff, Samuel W. and Durand, Frédo},
   year={2017},
   month=jul, pages={1–12} }

@misc{atd,
      title={Transcending the Limit of Local Window: Advanced Super-Resolution Transformer with Adaptive Token Dictionary}, 
      author={Leheng Zhang and Yawei Li and Xingyu Zhou and Xiaorui Zhao and Shuhang Gu},
      year={2024},
      eprint={2401.08209},
      archivePrefix={arXiv},
      primaryClass={cs.CV},
      url={https://arxiv.org/abs/2401.08209}, 
}

@article{pft,
  title={Progressive Focused Transformer for Single Image Super-Resolution},
  author={Long, Wei and Zhou, Xingyu and Zhang, Leheng and Gu, Shuhang},
  journal={arXiv preprint arXiv:2503.20337},
  year={2025}
}

@ARTICLE{contrast_enhancer,
  author={Cai, Jianrui and Gu, Shuhang and Zhang, Lei},
  journal={IEEE Transactions on Image Processing}, 
  title={Learning a Deep Single Image Contrast Enhancer from Multi-Exposure Images}, 
  year={2018},
  volume={27},
  number={4},
  pages={2049-2062},
  doi={10.1109/TIP.2018.2794218}}

@article{3dlut,
   title={Learning Image-adaptive 3D Lookup Tables for High Performance Photo Enhancement in Real-time},
   ISSN={1939-3539},
   url={http://dx.doi.org/10.1109/TPAMI.2020.3026740},
   DOI={10.1109/tpami.2020.3026740},
   journal={IEEE Transactions on Pattern Analysis and Machine Intelligence},
   publisher={Institute of Electrical and Electronics Engineers (IEEE)},
   author={Zeng, Hui and Cai, Jianrui and Li, Lida and Cao, Zisheng and Zhang, Lei},
   year={2020},
   pages={1–1} }

@misc{he2020conditionalsequentialmodulationefficient,
      title={Conditional Sequential Modulation for Efficient Global Image Retouching}, 
      author={Jingwen He and Yihao Liu and Yu Qiao and Chao Dong},
      year={2020},
      eprint={2009.10390},
      archivePrefix={arXiv},
      primaryClass={cs.CV},
      url={https://arxiv.org/abs/2009.10390}, 
}

@misc{ignatov2021learnedsmartphoneispmobile,
      title={Learned Smartphone ISP on Mobile NPUs with Deep Learning, Mobile AI 2021 Challenge: Report}, 
      author={Andrey Ignatov and Cheng-Ming Chiang and Hsien-Kai Kuo and Anastasia Sycheva and Radu Timofte and Min-Hung Chen and Man-Yu Lee and Yu-Syuan Xu and Yu Tseng and Shusong Xu and Jin Guo and Chao-Hung Chen and Ming-Chun Hsyu and Wen-Chia Tsai and Chao-Wei Chen and Grigory Malivenko and Minsu Kwon and Myungje Lee and Jaeyoon Yoo and Changbeom Kang and Shinjo Wang and Zheng Shaolong and Hao Dejun and Xie Fen and Feng Zhuang and Yipeng Ma and Jingyang Peng and Tao Wang and Fenglong Song and Chih-Chung Hsu and Kwan-Lin Chen and Mei-Hsuang Wu and Vishal Chudasama and Kalpesh Prajapati and Heena Patel and Anjali Sarvaiya and Kishor Upla and Kiran Raja and Raghavendra Ramachandra and Christoph Busch and Etienne de Stoutz},
      year={2021},
      eprint={2105.07809},
      archivePrefix={arXiv},
      primaryClass={eess.IV},
      url={https://arxiv.org/abs/2105.07809}, 
}

@article{simonyan2014very,
  title = {Very Deep Convolutional Networks for Large-Scale Image Recognition},
  author = {Simonyan, Karen and Zisserman, Andrew},
  journal = {arXiv preprint arXiv:1409.1556},
  year = {2014}
}

@ARTICLE{niqe,
  author={Zhang, Lin and Zhang, Lei and Bovik, Alan C.},
  journal={IEEE Transactions on Image Processing}, 
  title={A Feature-Enriched Completely Blind Image Quality Evaluator}, 
  year={2015},
  volume={24},
  number={8},
  pages={2579-2591},
  doi={10.1109/TIP.2015.2426416}}

@misc{wang2022exploringclipassessinglook,
      title={Exploring CLIP for Assessing the Look and Feel of Images}, 
      author={Jianyi Wang and Kelvin C. K. Chan and Chen Change Loy},
      year={2022},
      eprint={2207.12396},
      archivePrefix={arXiv},
      primaryClass={cs.CV},
      url={https://arxiv.org/abs/2207.12396}, 
}

@misc{yang2022maniqamultidimensionattentionnetwork,
      title={MANIQA: Multi-dimension Attention Network for No-Reference Image Quality Assessment}, 
      author={Sidi Yang and Tianhe Wu and Shuwei Shi and Shanshan Lao and Yuan Gong and Mingdeng Cao and Jiahao Wang and Yujiu Yang},
      year={2022},
      eprint={2204.08958},
      archivePrefix={arXiv},
      primaryClass={cs.CV},
      url={https://arxiv.org/abs/2204.08958}, 
}

@misc{chen2023topiqtopdownapproachsemantics,
      title={TOPIQ: A Top-down Approach from Semantics to Distortions for Image Quality Assessment}, 
      author={Chaofeng Chen and Jiadi Mo and Jingwen Hou and Haoning Wu and Liang Liao and Wenxiu Sun and Qiong Yan and Weisi Lin},
      year={2023},
      eprint={2308.03060},
      archivePrefix={arXiv},
      primaryClass={cs.CV},
      url={https://arxiv.org/abs/2308.03060}, 
}

@InProceedings{yan2025mobileie,
    author    = {Yan, Hailong and Li, Ao and Zhang, Xiangtao and Liu, Zhe and Shi, Zenglin and Zhu, Ce and Zhang, Le},
    title     = {MobileIE: An Extremely Lightweight and Effective ConvNet for Real-Time Image Enhancement on Mobile Devices},
    booktitle = {Proceedings of the IEEE/CVF International Conference on Computer Vision},
    month     = {October},
    year      = {2025},
}

@misc{lou2026perceivingbettermomentscover,
      title={Perceiving Better Moments: Cover Frame Reselection and Enhancement for Live Photos with the Live2K Dataset}, 
      author={Junyu Lou and Kai Chen and Weiyi You and Hui Zeng and Lei Zhang and Shuhang Gu},
      year={2026},
      eprint={2607.04151},
      archivePrefix={arXiv},
      primaryClass={cs.CV},
      url={https://arxiv.org/abs/2607.04151}, 
}

@misc{chen2025taskawareimagesignalprocessor,
      title={Task-Aware Image Signal Processor for Advanced Visual Perception}, 
      author={Kai Chen and Jin Xiao and Leheng Zhang and Kexuan Shi and Shuhang Gu},
      year={2025},
      eprint={2509.13762},
      archivePrefix={arXiv},
      primaryClass={cs.CV},
      url={https://arxiv.org/abs/2509.13762}, 
}

@misc{cui2025rawadapteradaptingpretrainedvisual,
      title={RAW-Adapter: Adapting Pre-trained Visual Model to Camera RAW Images and A Benchmark}, 
      author={Ziteng Cui and Jianfei Yang and Tatsuya Harada},
      year={2025},
      eprint={2503.17027},
      archivePrefix={arXiv},
      primaryClass={cs.CV},
      url={https://arxiv.org/abs/2503.17027}, 
}

@INPROCEEDINGS{raod,
  author={Xu, Ruikang and Chen, Chang and Peng, Jingyang and Li, Cheng and Huang, Yibin and Song, Fenglong and Yan, Youliang and Xiong, Zhiwei},
  booktitle={2023 IEEE/CVF Conference on Computer Vision and Pattern Recognition (CVPR)}, 
  title={Toward RAW Object Detection: A New Benchmark and A New Model}, 
  year={2023},
  volume={},
  number={},
  pages={13384-13393},
  doi={10.1109/CVPR52729.2023.01286}}

@misc{wang2024adaptiveisplearningadaptiveimage,
      title={AdaptiveISP: Learning an Adaptive Image Signal Processor for Object Detection}, 
      author={Yujin Wang and Tianyi Xu and Fan Zhang and Tianfan Xue and Jinwei Gu},
      year={2024},
      eprint={2410.22939},
      archivePrefix={arXiv},
      primaryClass={cs.CV},
      url={https://arxiv.org/abs/2410.22939}, 
}
}
\clearpage
\setcounter{page}{1}
\setcounter{section}{0}
\renewcommand{\thesection}{\Alph{section}}
\maketitlesupplementary
\section{More visualization of LC-ISP}
\begin{figure*}[t]
    \centering
    \includegraphics[width=0.95\textwidth,height=0.86\textheight,keepaspectratio]{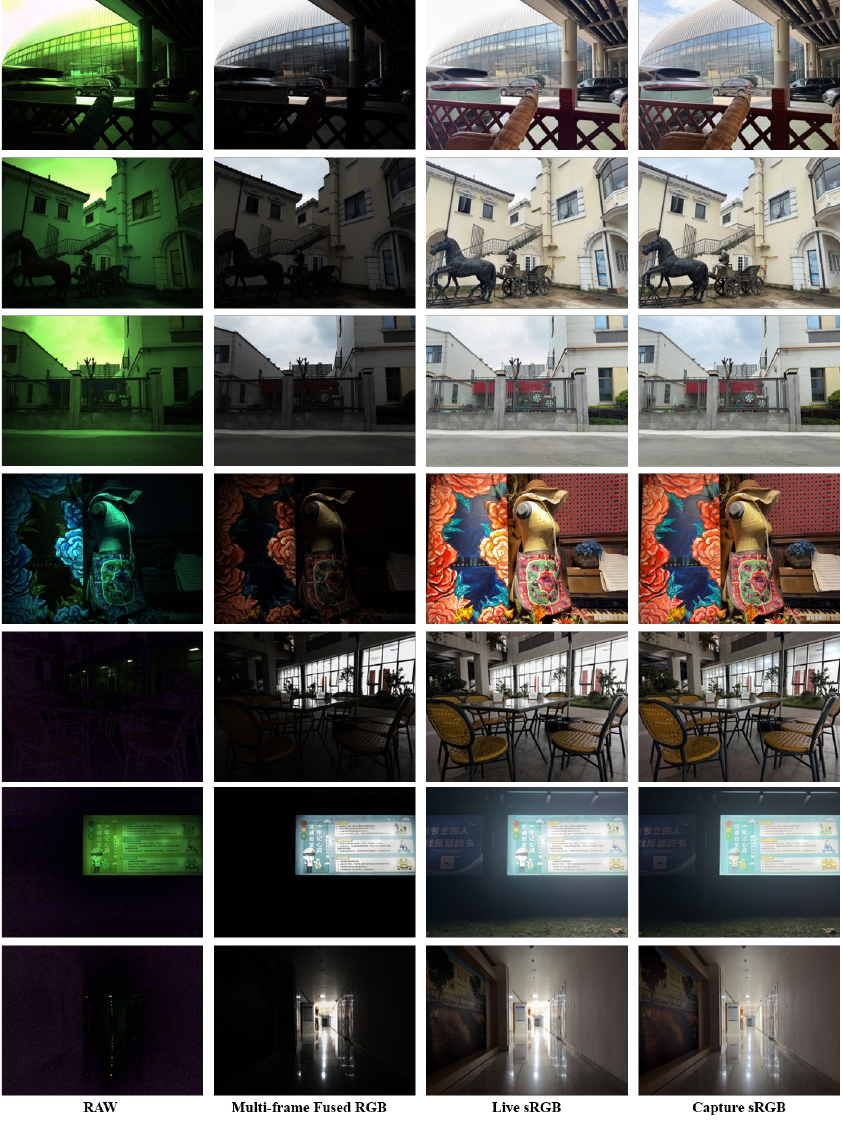}
    \caption{Visualization of diverse scenes from our LC-ISP dataset. The first four rows depict daytime scenes, while the last three rows illustrate nighttime scenes.}
    \label{sup_dataset}
\end{figure*}

In this section, we provide additional visual examples of LC-ISP on our dataset, as shown in Figure.~\ref{sup_dataset}. The examples cover diverse real-world scenarios, including daytime and nighttime, indoor and outdoor scenes, as well as challenging high dynamic range (HDR) cases.

\section{Additional comparisons with other methods}
\begin{figure*}[t]
    \centering
    \includegraphics[width=0.95\textwidth]{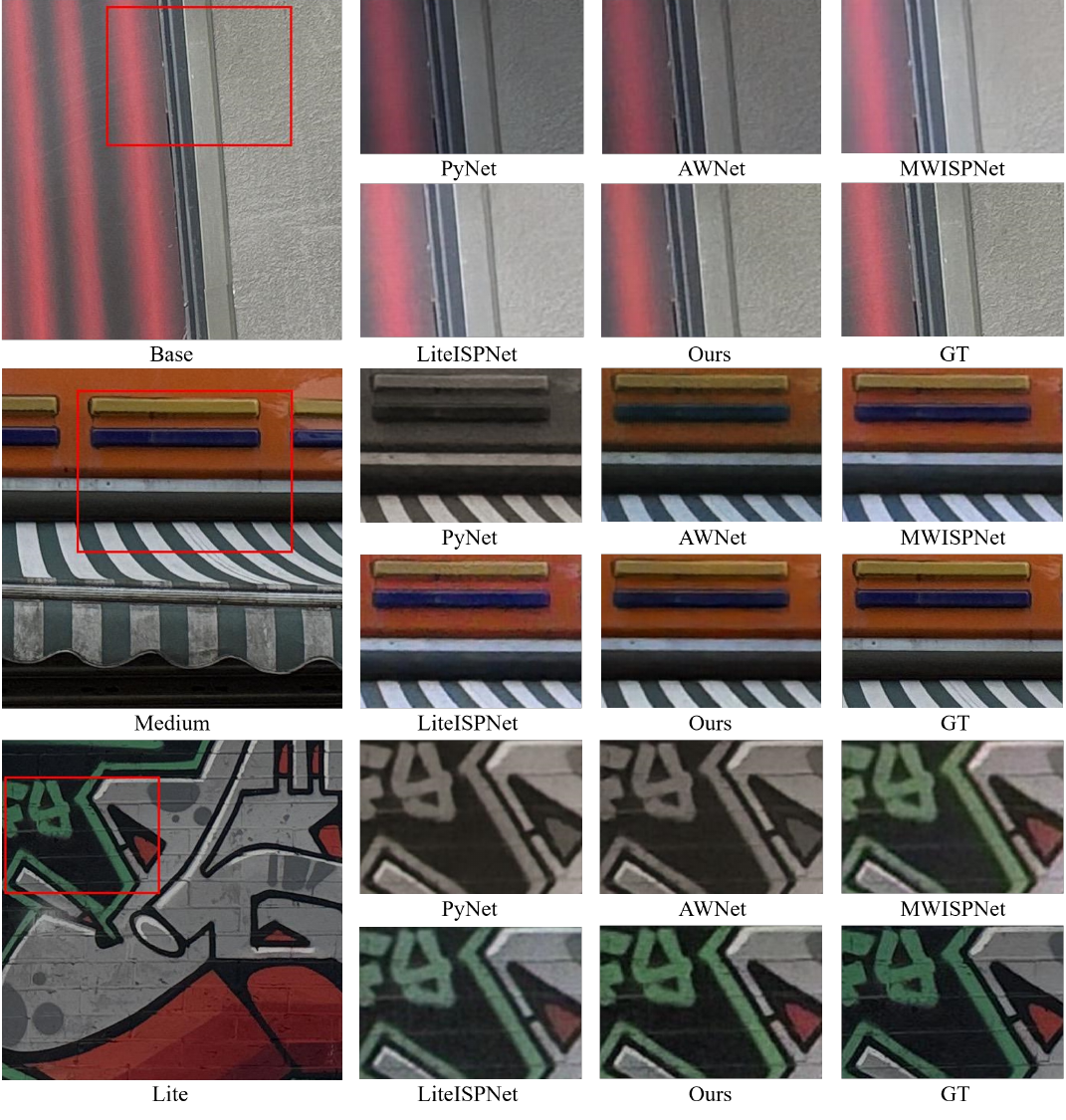}
    \caption{Visual comparison of competing methods. Please zoom in for more details.}
    \label{sup_compare0}
\end{figure*}

\begin{figure*}[t]
    \centering
    \includegraphics[width=0.95\textwidth]{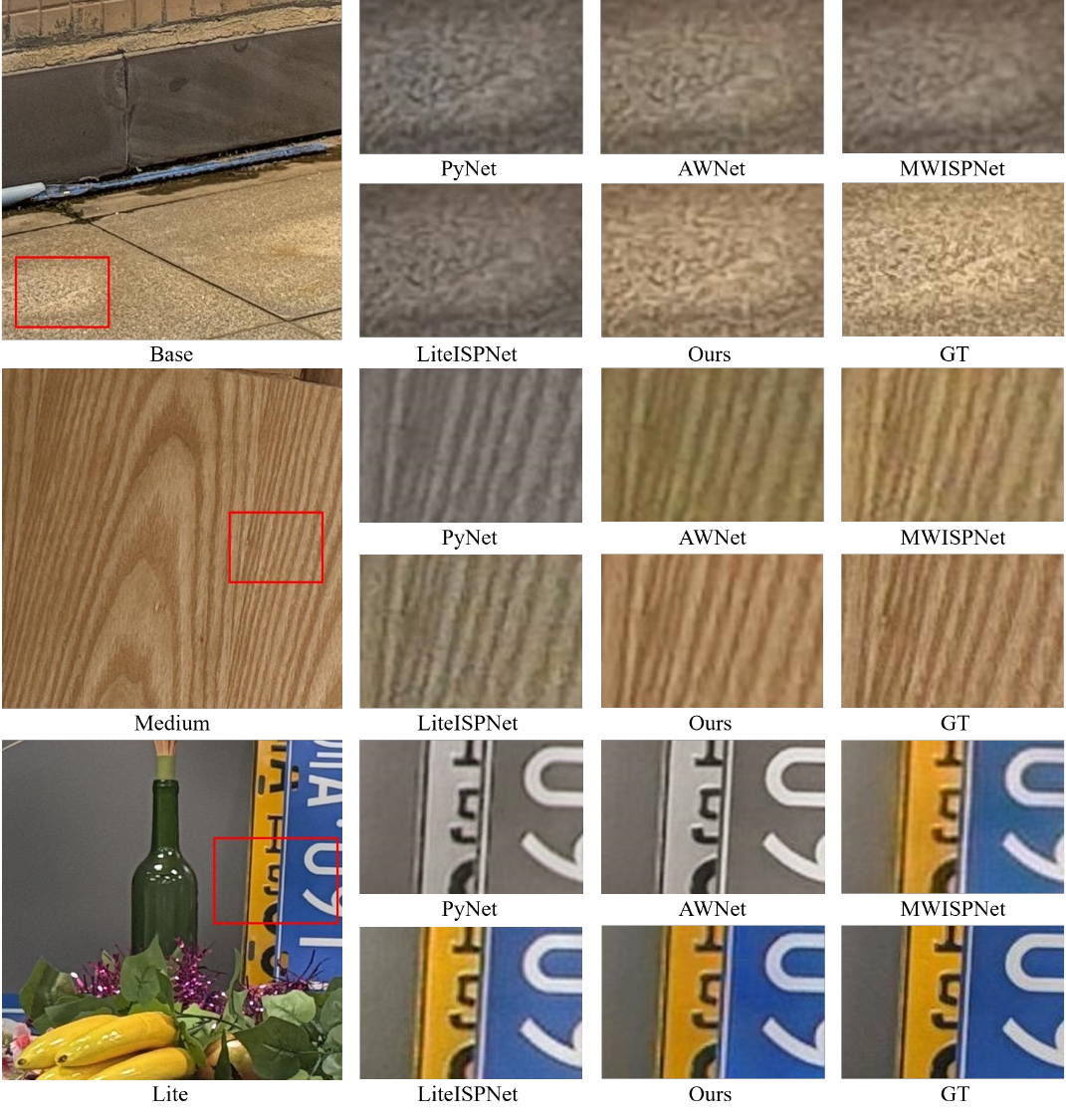}
    \caption{Visual comparison of competing methods. Please zoom in for more details.}
    \label{sup_compare}
\end{figure*}

To better demonstrate the superiority of our Dual-Stage ISP, we provide additional visual comparisons under three model-size settings, namely base, medium, and lite, against other single-frame ISP methods. As shown in Figure~\ref{sup_compare0} and Figure~\ref{sup_compare}, our approach consistently outperforms all competing single-frame ISP algorithms across different model capacities. The advantage is particularly pronounced in the lite setting, where the computational budget is most restricted, highlighting the effectiveness of our design under strict resource constraints.

\end{document}